\documentclass[onecolumn]{article} 

\usepackage{multicol}  
\usepackage[letterpaper, top=0.6in, bottom=0.5in, inner=0.75in, outer=0.75in, headsep=0.3in, footskip=0.3in]{geometry}

\usepackage{titlesec}
\titleformat{\section}{\large\bfseries}{\thesection}{1em}{}
\titleformat{\subsection}{\normalsize\bfseries}{\thesubsection}{1em}{}

\def\huggingface{\raisebox{-1.5pt}{\includegraphics[height=1.05em]{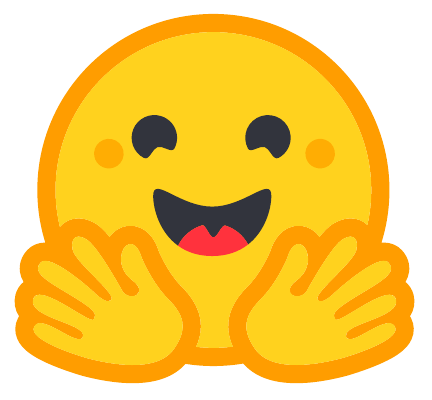}}}

\usepackage{tabularx} 

\usepackage{graphicx}         
\usepackage{amsmath, amssymb} 
\usepackage{hyperref}          
\usepackage{natbib}          
\usepackage{geometry}         
\usepackage{listings}
\usepackage{pgfplots}
\usepackage{amssymb} 
\usepackage{pifont}  
\usepgfplotslibrary{fillbetween}
\usepackage{algorithm}
\usepackage{algorithmic}
\usepackage{amsmath} 
\usepackage{appendix}
\usepackage{booktabs} 
\usepackage{caption}  
\usepackage{array} 
\title{\textbf{Jan-nano Technical Report}}

\author{
    Alan Dao (Gia Tuan Dao)\textsuperscript{1}, Dinh Bach Vu\textsuperscript{1} \\ \\[0.75em]
    Menlo Research \\ [0.5em]
    \texttt{alan@menlo.ai, bach@menlo.ai} \\ [0.5em]
\begin{tabular}{rl}
\huggingface & \url{https://huggingface.co/Menlo/Jan-nano} \\
\huggingface & \url{https://huggingface.co/Menlo/Jan-nano-gguf} \\
\huggingface & \url{https://huggingface.co/Menlo/Jan-nano-128k} \\
\huggingface & \url{https://huggingface.co/Menlo/Jan-nano-128k-gguf} \\
\end{tabular}
}
\usepackage{datetime}
\date{\formatdate{1}{7}{2025}} 

\usepackage{fancyhdr} 
\begin{document}

\pagestyle{fancy}
\fancyhf{}
\fancyhead[L]{\includegraphics[height=2.7em]{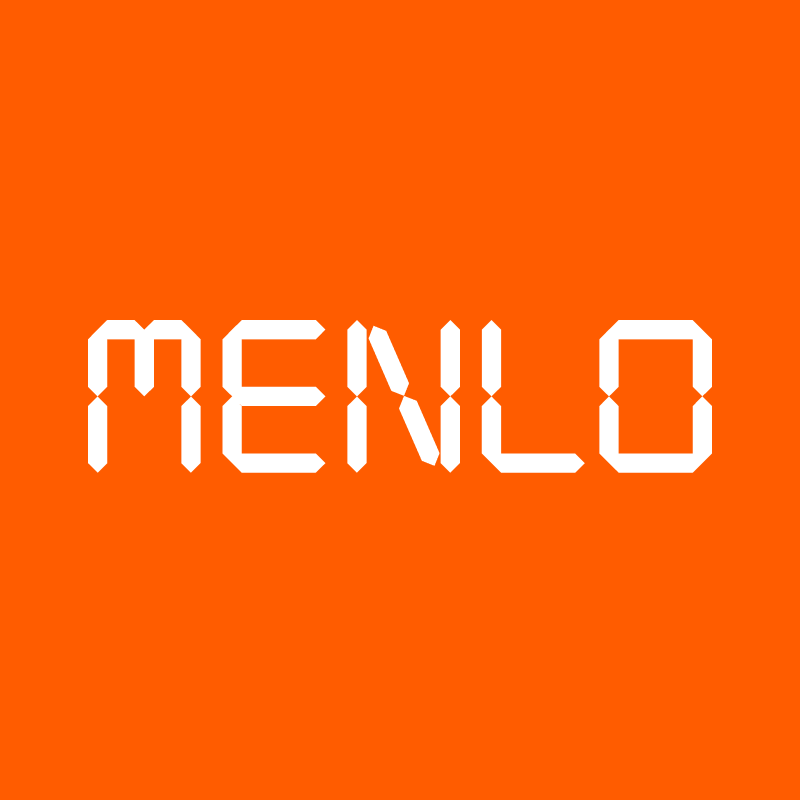}}
\fancyhead[R]{\today}
\fancyfoot[C]{\thepage}
\renewcommand{\headrulewidth}{0.5pt}
\fancypagestyle{plain}{\pagestyle{fancy}}

\maketitle
\begin{abstract}
Most language models face a fundamental tradeoff where powerful capabilities require substantial computational resources. We shatter this constraint with Jan-nano, a 4B parameter language model that redefines efficiency through radical specialization: instead of trying to know everything, it masters the art of finding anything instantly. Fine-tuned from Qwen3-4B using our novel multi-stage Reinforcement Learning with Verifiable Rewards (RLVR) system that \textbf{completely eliminates reliance on next token prediction training (SFT)}, Jan-nano \textbf{achieves 83.2\% on SimpleQA benchmark with MCP integration} while running on consumer hardware. With 128K context length, Jan-nano proves that intelligence isn't about scale, it's about strategy.
\noindent 
\end{abstract}

\section{Introduction}
\label{sec:introduction}

The development of large language models (LLMs) has traditionally been characterized by a fundamental trade-off between performance and computational demand. Achieving state-of-the-art capabilities often necessitates models with tens or hundreds of billions of parameters, rendering them impractical for deployment on consumer-grade hardware and limiting their accessibility. This constraint forces a compromise between a model's reasoning and knowledge capacity and its operational efficiency.

To mitigate this, a prominent line of research has focused on augmenting LLMs with external information sources through Retrieval-Augmented Generation (RAG)~\cite{lewis2021retrievalaugmentedgenerationknowledgeintensivenlp}. Recent approaches, such as Search-R1~\cite{jin2025searchr1trainingllmsreason}, ReSearch~\cite{chen2025researchlearningreasonsearch}, and DeepResearcher~\cite{zheng2025deepresearcherscalingdeepresearch}, have demonstrated the efficacy of this paradigm. These methods typically leverage reinforcement learning techniques like Group Relative Policy Optimization (GRPO) on well-established 7B-parameter models to enhance performance on complex, multi-hop question-answering benchmarks like HotpotQA~\cite{yang2018hotpotqadatasetdiverseexplainable} and 2WikiMultiHopQA~\cite{ho-etal-2020-constructing}. However, these efforts have predominantly targeted larger model architectures, leaving the potential of smaller, highly efficient sub-5B parameter models for sophisticated tool-augmented reasoning largely under-explored.

In this report, we introduce \textbf{Jan-nano}, a specialized 4B-parameter model designed to address this gap. Jan-nano is built on a distinct design philosophy: we prioritize proficient information retrieval and tool utilization over the conventional approach of storing encyclopedic knowledge within the model's parameters. Our core principle is that by engineering a model to excel at finding information, we can achieve competitive performance on knowledge-intensive tasks while maintaining a computational footprint suitable for local deployment. To this end, Jan-nano is fine-tuned from the state-of-the-art Qwen3-4B base model~\cite{yang2025qwen3technicalreport}.
Our primary contribution is a novel multi-stage training framework, RLVR, which diverges from conventional SFT-based fine-tuning. We leverage the MuSiQue dataset~\cite{trivedi2022musiquemultihopquestionssinglehop} in a controlled environment with a local RAG server that simulates real-world search engine interactions. The training process employs Direct Alignment from Preference Optimization (DAPO)~\cite{yu2025dapoopensourcellmreinforcement} and incorporates a regularization technique to penalize superfluous generation steps, thereby promoting concise and direct tool invocation. Our evaluation on the SimpleQA benchmark~\cite{wei2024measuringshortformfactualitylarge} demonstrates that this strategic design allows Jan-nano to achieve strong results, showing that highly specialized, smaller models can effectively overcome traditional capability-efficiency constraints in tool-augmented tasks.

\section{Performance Benchmarks}
\label{sec:perf_benchmarks}

The benchmark results include both models with and without MCP (Model Context Protocol) integration. For our MCP evaluation, we used mcp-server-serper\footnote{\url{https://github.com/marcopesani/mcp-server-serper}} which provides \texttt{google\_search} and \texttt{scrape} tools.

\begin{table}[H]
    \centering
    \begin{tabular}{lcc}
    \toprule
    \textbf{Model} & \textbf{SimpleQA} & \textbf{Parameters} \\
    \midrule
    OpenAI o1 & 42.6\% & Unknown \\
    Grok 3 & 44.6\% & Unknown \\
    o3 & 49.4\% & Unknown \\
    Claude-3.7-Sonnet & 50.0\% & Unknown \\
    Gemini-2.5 Pro & 52.9\% & Unknown \\
    Qwen3-4B (Baseline) (MCP) & 59.2\% & 4B \\
    ChatGPT-4.5 & 62.5\% & Unknown \\
    DeepSeek-671B (OpenRouter) (MCP) & 78.2\% & 671B \\
    Jan-nano (MCP) & 80.7\% & 4B \\
    Jan-nano-128k (MCP) & \textbf{83.2\%} & 4B \\
    \bottomrule
    \end{tabular}
    \caption{SimpleQA benchmark~\cite{wei2024measuringshortformfactualitylarge} results.}
    \label{tab:benchmarks}
\end{table}

Jan-nano achieves competitive performance at 83.2\% accuracy with 4B parameters, representing a notable 24 percentage point improvement over the Qwen3-4B baseline with MCP (59.2\%). When compared to DeepSeek-671B with MCP (78.2\%), our model demonstrates improved parameter efficiency, though we note that different MCP implementations and evaluation setups may affect direct comparisons. The Jan-nano 128K context variant shows a modest 2.5 percentage point improvement over the standard variant, suggesting benefits from extended context length for this benchmark.
Our experimental setup deliberately employed a minimal MCP configuration to establish a conservative baseline. We utilized sample code from LangGraph connected to an MCP server, specifically the mcp-server-serper\footnote{\url{https://github.com/marcopesani/mcp-server-serper}}, following the same integration approach used in standard chat applications. Notably, we maintained a bare-minimum setup without addressing known bugs in the MCP implementation, allowing some operations to fail naturally. While we developed a forked version that resolved certain MCP issues and could potentially achieve above 85\% accuracy, we opted for the more conservative 83.2\% result to avoid inflated performance claims. This conservative approach reflects our understanding that model performance is inherently tied to the quality of tool implementations - superior MCP setups with features like pagination, meaningful error handling, and robust scrapers that bypass robots.txt restrictions could yield significantly higher scores. Our evaluation demonstrates typical failure patterns where the model reads until encountering familiar mid-training concepts and terminates prematurely, rather than exhaustively processing available information, which would likely cause context timeouts with conventional scraping approaches.

\section{Data Preparation}
\label{sec:data_preparation}

We use the MuSiQue-Ans dataset~\cite{trivedi2022musiquemultihopquestionssinglehop}, a multihop QA dataset that systematically selects composable pairs of single-hop questions connected through multihop reasoning. This dataset was selected for its methodology of progressive difficulty increase (2-hop, 3-hop, 4-hop questions) and the availability of supporting paragraphs. A subset of 10,325 samples was selected: 7,000 2-hop (67.8\%), 2,150 3-hop (20.8\%), and 1,175 4-hop questions (11.4\%). 

\section{Training Methodology}
\label{sec:methodology}

Our training methodology focuses on developing efficient tool usage capabilities while preserving model autonomy. Starting from Qwen3-4B~\cite{yang2025qwen3technicalreport}, we establish a controlled training environment using the MuSiQue dataset with a local RAG server that simulates real search engine functionality.

We employed a \textbf{multi-stage RLVR system that completely eliminates reliance on next token prediction training (SFT)}. This approach leverages verifiable reward signals to guide learning without requiring massive datasets. The training follows a three-stage progression using DAPO (Direct Alignment from Preference Optimization)~\cite{yu2025dapoopensourcellmreinforcement} with force non-thinking to prevent overthinking behaviors: (1) tool usage foundation where the model learns basic tool mechanics, (2) answer quality focus where rewards shift to prioritize correctness, and (3) context extension from 8K to 40K tokens.

\begin{figure*}[p]
    \centering
    \includegraphics[width=1.0\textwidth]{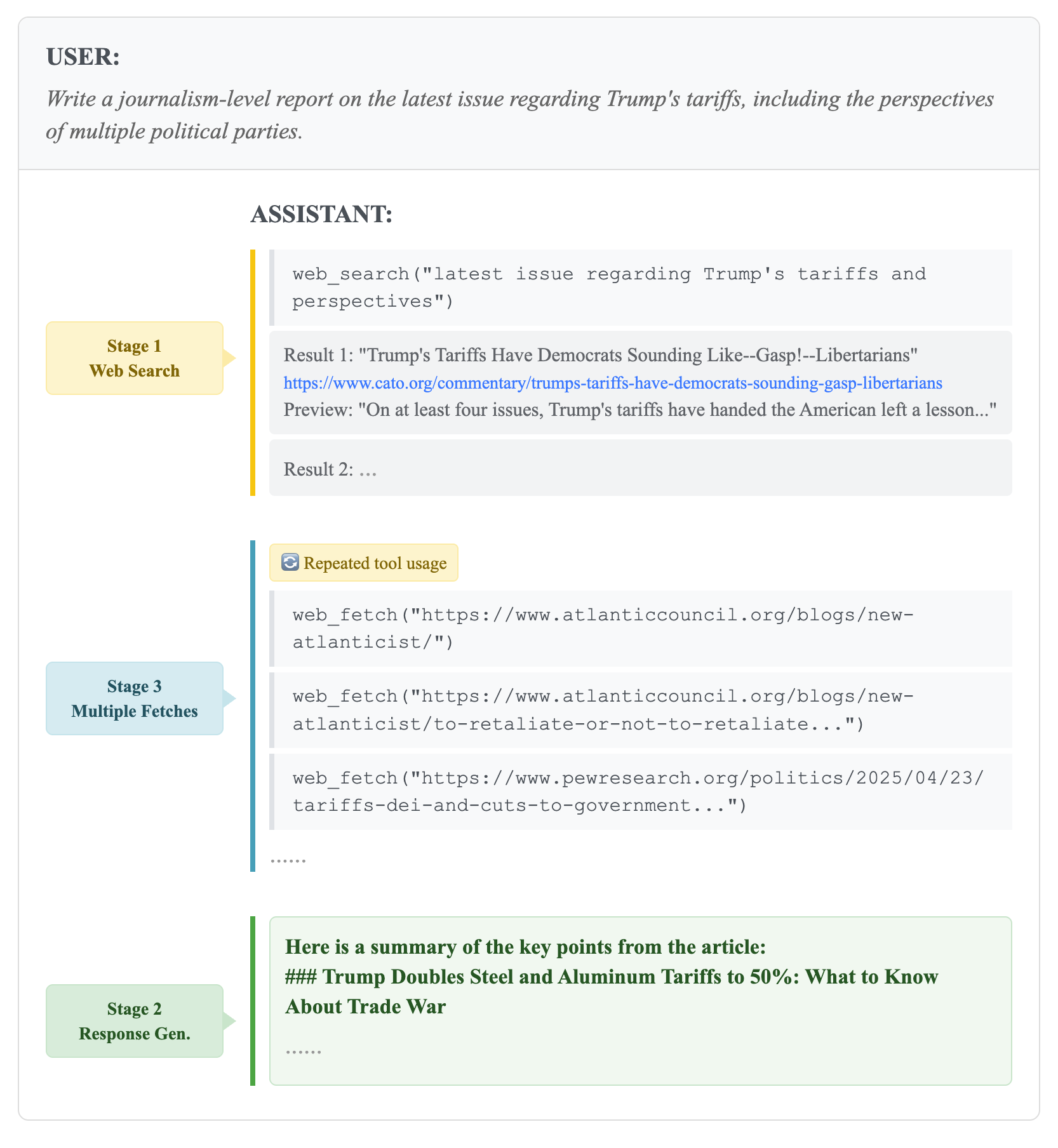}
    \caption{Visualization of focus during different training stages.}
    \label{fig:methodology}
\end{figure*}

Throughout training, we maintain a JSON-in-XML format for tool interactions. Tools are invoked using \texttt{<tool>
\{"name": "tool\_name", "args": \{params\}\}</tool>} tags, results appear in \texttt{<result>content</result>} tags, and final answers use \texttt{<answer>content</answer>} tags. This structured XML format facilitates reliable parsing and comparison during RL training while enabling the model to develop autonomous reasoning patterns with minimal system prompts as illustrated in Figure \ref{fig:methodology}.

\subsection{Local RAG Server to Mimic Search Engine and Web Scrape}
We developed a local RAG server to simulate search engine and web scraping functionality during training, providing controlled access to the MuSiQue corpus. The implementation employs a two-stage retrieval pipeline that combines dense retrieval with cross-encoder reranking to produce high-quality search results that mirror real-world search behavior.

The indexing process utilizes E5-base-v2\footnote{\url{https://huggingface.co/intfloat/e5-base-v2}} encoder to generate dense embeddings for all corpus documents, which are then indexed using FAISS\footnote{\url{https://github.com/facebookresearch/faiss}} with GPU acceleration for efficient similarity search. Each document maintains a structured format with title and text content stored in JSONL format, enabling rapid access and retrieval.

The server exposes two primary tools that mirror typical search engine functionality: \texttt{websearch(query)} returns up to 10 relevant documents with 150-character previews, simulating search result snippets, while \texttt{scrape(doc\_id)} retrieves the full content of a specific document using its unique identifier. This design encourages the model to learn efficient search strategies by first examining previews before selectively accessing full documents.

Server configuration parameters were optimized for training efficiency: initial retrieval returns top-15 documents which are then reranked using cross-encoder model ms-marco-MiniLM-L12-v2\footnote{\url{https://huggingface.co/cross-encoder/ms-marco-MiniLM-L12-v2}} to select the final top-10 results. This setup provides a controlled environment for training tool usage capabilities while maintaining reproducibility across training runs.

\subsection{Stage 1: Tool Usage Foundation (8K Context)}
The model learns basic tool mechanics and interaction patterns with 8K context length. The reward function balances multiple objectives: correct answer, successful tool execution, format adherence, and XML structure compliance. This stage establishes fundamental tool calling behaviors and response formatting.

\subsection{Stage 2: Answer Quality Focus (8K Context)}
This stage shifts the emphasis to answer correctness. We drop tool execution and format rewards, focusing the model to increase accuracy. The reward function prioritizes correct answers while maintaining minimal XML structure guidance.

\subsection{Stage 3: Context Extension (40K Context)}
This final training stage extends the model's capability to handle longer contexts by increasing the context window from 8K to 40K tokens. The reward function only emphasizes correctness and XML structure similar to the previous stage, ensuring the model maintains high-quality responses while adapting to extended context lengths.

\section{Key Findings}
\label{sec:key_findings}

\subsection{Overthinking Problem in Language Models}
After three iterations of model development, we systematically compared Think versus No-Think modes across configurations. Analysis revealed that larger models exhibit overthinking behaviors that degrade search performance through unnecessary additional filter parameters and over-complex search approaches. The 4B model uses direct search without filtering constraints, successfully retrieving accurate information, while the 8B model applies temporal filtering that eliminates relevant results, leading to hallucinated responses (see Appendix~\ref{appendix:a1} for complete conversation logs demonstrating these behaviors). These overthinking behaviors are more pronounced in larger models, with 8B parameter variants exhibiting increased over-planning and analysis paralysis compared to 4B models, as quantified in our systematic evaluation:

\vspace{0.5em}

\begin{table}[H]
\centering
\begin{tabular}{lcc}
\toprule
\textbf{Model Size} & \textbf{Mode} & \textbf{SimpleQA Score} \\
\midrule
4B & Think & 71.0\% \\
8B & Think & 62.2\% \\
\bottomrule
\end{tabular}
\caption{Performance comparison of Think mode on 10\% SimpleQA subset using Serper search tool.}
\label{tab:think-vs-nothink}
\end{table}

Analysis across model configurations reveals that disabling the thinking mechanism provides substantial speed benefits while maintaining competitive accuracy. Through comparative evaluation of same model sizes with and without thinking capabilities, performance differences remain relatively modest, yet no-think variants deliver significantly faster response times. This speed advantage, combined with improved user experience from reduced latency, motivated our adoption of force non-thinking training in subsequent versions.

\subsection{Agentic Configuration Assessment}

To evaluate the Jan-nano model, we employed different workflow paradigms to assess performance across varying execution approaches. We examined two primary categories: traditional structured workflows where the LLM operates within predefined code paths, and agentic workflows where the language model autonomously directs its own actions based on environmental feedback. In agentic workflows, the model maintains control over decision-making processes, tool selection, and execution flow.

Since we aimed to develop Jan-nano with autonomous reasoning and decision-making capabilities, we focused our evaluation on agentic workflows that would best demonstrate these characteristics. We initially tested two implementations: CodeAgent from smolagents\footnote{\url{https://github.com/huggingface/smolagents}} and LangGraph-based ReAct\footnote{\url{https://github.com/langchain-ai/langgraph}}.Initial evaluation using the CodeAgent framework revealed performance limitations due to template format mismatch. Jan-nano was trained with JSON tool call templates but CodeAgent requires Python code generation—a format the model was not optimized for. Subsequent evaluation with the LangGraph ReAct implementation, implemented using MCP (Model Context Protocol) servers, which uses JSON-formatted tool calls matching Jan-nano's training template, demonstrated improved performance (see Appendix~\ref{appendix:a2}).

Through this evaluation process, we recognized that Model Context Protocol (MCP) tooling provides the most flexible framework to replicate authentic agentic application user experiences. Unlike traditional evaluation frameworks that impose rigid constraints on tool integration and interaction patterns, MCP enables seamless integration of diverse tools and services, allowing the model to interact with real-world systems in ways that mirror actual deployment scenarios. This flexibility proved essential for comprehensive performance assessment, as MCP's standardized protocol permits dynamic tool discovery and utilization—capabilities that are fundamental to practical agentic applications.
Consequently, we adopted MCP as our primary evaluation framework, which enabled comprehensive assessment of Jan-nano's agentic capabilities while providing the operational flexibility necessary for realistic performance evaluation.
\section{Limitations and Future Work}
\label{sec:limitations_and_future_work}

While Jan-nano demonstrates a significant step forward in creating efficient, tool-specialized models, we acknowledge several limitations that provide clear avenues for future research. Currently, the model's final performance of 83.2\% on the SimpleQA benchmark indicates a gap between successful tool invocation and final answer synthesis. We have observed instances where the model retrieves the correct document but fails to accurately reason over the provided context, a limitation likely stemming from the inherent generative constraints of a 4B parameter architecture. Furthermore, its training has been narrowly focused on fact-based question answering, and its capabilities on open-ended conversational or instructional tasks remain un-assessed. Building on our current framework, future work will aim to reduce hallucination in open-ended research-oriented tasks, enabling the model to generate more trustworthy and well-supported reports. To achieve this, we plan to refine the RLVR framework by introducing a more nuanced reward model that not only validates factual accuracy but also incentivizes proper citation of retrieved sources. Additionally, we will expand Jan-nano's capabilities by integrating a more diverse and complex set of tools, such as a \textit{Calculator} for mathematical reasoning and a \textit{Code Interpreter} for computational tasks, further grounding its responses in verifiable external processes and enhancing its utility as a reliable research assistant.

\section{Conclusion}
\label{sec:conclusion}

In this report, we have presented Jan-nano, a 4B parameter language model that demonstrates a novel approach to resolving the fundamental trade-off between model capability and computational efficiency. Our primary contribution is the multi-stage RLVR training system, which completely eschews traditional SFT in favor of a reinforcement learning paradigm guided by verifiable task outcomes. By leveraging DAPO and a "force non-thinking" regularization strategy, Jan-nano is explicitly optimized for efficient tool usage, avoiding the "overthinking" behaviors that we observed to degrade the performance of larger models.

The efficacy of this approach is demonstrated by Jan-nano's 83.2\% accuracy on the SimpleQA benchmark with MCP integration, where it significantly outperforms its baseline and remains highly competitive with models orders of magnitude larger. Ultimately, Jan-nano serves as a proof of concept that intelligence is not merely a function of parameter count, but of strategic design. It validates a promising path forward for the development of highly capable, efficient, and specialized language models that can operate effectively on consumer-grade hardware, redefining what is possible for local AI.

\bibliographystyle{plainnat} 
\bibliography{main} 


\clearpage
\appendix
\onecolumn
\appendix
\section{Conversation Logs Demonstrating Overthinking Behaviors}
\label{sec:appendix}

\subsection{Comparative Analysis of Year Filter Constraints Leading to Hallucination}\label{appendix:a1}

We present two representative examples demonstrating how the 8B model's use of additional parameters systematically leads to search failures and hallucinated answers, while the 4B model's direct approach succeeds in both cases.

\begin{table}[H]
\centering
\scriptsize
\setlength{\tabcolsep}{3pt}
\renewcommand{\arraystretch}{1.1}
\begin{tabular}{p{0.12\textwidth}>{\raggedright\arraybackslash}p{0.42\textwidth}>{\raggedright\arraybackslash}p{0.42\textwidth}}
\toprule
\textbf{Query} & \textbf{4B Model (Successful Trace)} & \textbf{8B Model (Failing Trace)} \\
\midrule
\textbf{Example 1:} \textit{"What is the name of the Dutch businessman and convicted drug trafficker who trained two of the hijackers of the planes used on 9/11?"} & 
\textbf{ASSISTANT:} Here are the facts I know and the plan of action that I will follow:

\texttt{Facts given in the task}: The task provides the question...

\texttt{Facts to look up}: - Name of the Dutch businessman... - Confirmation of the individual's role...

\texttt{Plan}: 1. Perform a web search... 2. Use the search results to pinpoint the name...

\rule{0pt}{3ex} \textbf{ASSISTANT:} \textbf{Thought:} I will use the web\_search tool to find information...

\textbf{Action:} \texttt{search\_result = web\_search(query="Dutch businessman convicted drug trafficker trained 9/11 hijackers")}

\rule{0pt}{3ex} \textbf{ASSISTANT:} \textbf{Observation:} Search Results: 0. [Rudi Dekkers - Wikipedia]... Rudi Dekkers (July 27, 1956 – April 11, 2024) was a Dutch businessman and convicted drug trafficker. ... trained 9/11 terrorists... 1. [Head of 9/11 Hijackers' Flight School Faces Drug Running Charges]...

\rule{0pt}{3ex} \textbf{ASSISTANT:} The Dutch businessman and convicted drug trafficker linked to training two of the 9/11 hijackers is \textbf{Rudi Dekkers}. He was a flight school owner who inadvertently trained terrorists and faced drug trafficking charges.

\rule{0pt}{3ex} \textbf{ASSISTANT:} \textbf{Final Answer:} \texttt{Rudi Dekkers} &

\textbf{ASSISTANT:} Here are the facts I know and the plan of action that I will follow:

\texttt{Facts to look up}: - The identity of the Dutch businessman...

\texttt{Plan}: 1. Use \texttt{web\_search} to find information...

\rule{0pt}{3ex} \textbf{ASSISTANT:} \textbf{Thought:} I will start by searching for the Dutch businessman...

\textbf{Action:} \texttt{result = web\_search(query="Dutch businessman convicted drug trafficking 9/11", filter\_year=2001)}

\rule{0pt}{3ex} \textbf{ASSISTANT:} \textbf{Observation:} Search Results: 0. [About The Bin Laden Family]... members of the Shakarshi family, whose name is linked to a money-laundering scandal and drug trafficking... 1. [The threat of Russian Organized Crime]...

\rule{0pt}{3ex} \textbf{ASSISTANT:} \textbf{Thought:} The initial search did not yield specific results. I need to adjust the query...

\textbf{Action:} \texttt{result = web\_search(query="Dutch nationals 9/11 hijackers drug trafficking", filter\_year=2001)}

\rule{0pt}{3ex} \textbf{ASSISTANT:} \textbf{Observation:} Search Results: [A list of irrelevant search results from 2001 about terrorism, the Patriot Act, and the World Trade Center attacks.]

\rule{0pt}{3ex} \textbf{ASSISTANT:} (The model continues this pattern of failed searches, always keeping the \texttt{filter\_year=2001} constraint, before eventually giving up.)

\rule{0pt}{3ex} \textbf{ASSISTANT:} \textbf{Final Answer:} \texttt{Randy Weaver} \\
\midrule
\textbf{Example 2:} \textit{"Who was the production manager of ThunderCats, the science-fantasy animated television series that was released in 1985?"} & 
\textbf{ASSISTANT:} Here are the facts I know and the plan of action that I will follow to solve the task:

\texttt{Facts given in the task}: ThunderCats is a science-fantasy... The series was released in 1985.

\texttt{Facts to look up}: Production manager of ThunderCats (1985)... Use \texttt{web\_search(...)}

\rule{0pt}{3ex} \textbf{ASSISTANT:} \textbf{Thought:} I will use the web\_search tool to find information about the production manager of ThunderCats from 1985.

\textbf{Action:} \texttt{production\_info = web\_search(query="ThunderCats 1985 production manager")}

\rule{0pt}{3ex} \textbf{ASSISTANT:} \textbf{Observation:} Search Results: 0. [Masaki Iizuka - ThunderCats wiki - Fandom] Masaki Iizuka is known producer and production manager. For the shows ThunderCats (1985 original series)... 1. [Thundercats (TV Series 1985–1989) - Full cast \& crew - IMDb]...

\rule{0pt}{3ex} \textbf{ASSISTANT:} The production manager of \textit{ThunderCats} (1985) is \textbf{Masaki Iizuka}, as confirmed by multiple sources including Fandom, IMDb, and archived articles.

\rule{0pt}{3ex} \textbf{ASSISTANT:} \textbf{Final Answer:} \texttt{Masaki Iizuka} &

\textbf{ASSISTANT:} Here are the facts I know and the plan of action that I will follow to solve the task:

\texttt{Facts to look up}: Production team details for ThunderCats (1985) Tool: \texttt{web\_search(..., filter\_year=1985)}

\texttt{Plan}: Use \texttt{web\_search} to find "production manager of ThunderCats 1985" and filter results to 1985.

\rule{0pt}{3ex} \textbf{ASSISTANT:} \textbf{Thought:} I will start by searching for the production manager of ThunderCats using a web search filtered to 1985 to ensure relevance.

\textbf{Action:} \texttt{search\_results = web\_search(query="production manager of ThunderCats 1985", filter\_year=1985)}

\rule{0pt}{3ex} \textbf{ASSISTANT:} \textbf{Observation:} Search Results: 0. [Television/RadloAge] ... Starting September 1985, a brand new 90-minute concept! THE PAW-PAWS... presented by HANNA-BARBERA'S. [Other irrelevant results from 1985 newspapers and magazines.]

\rule{0pt}{3ex} \textbf{ASSISTANT:} \textbf{Thought:} The initial search did not yield direct results... I will refine the search by targeting the studio (Hanna-Barbera)...

\textbf{Action:} \texttt{search\_results\_hanna = web\_search(query="ThunderCats 1985 production team Hanna-Barbera", filter\_year=1985)}

\rule{0pt}{3ex} \textbf{ASSISTANT:} \textbf{Observation:} Search Results: [Another list of irrelevant search results from 1985, mentioning Hanna-Barbera but not ThunderCats' production team.]

\rule{0pt}{3ex} \textbf{ASSISTANT:} \textbf{Thought:} Since "production manager" might not be the exact title used, and no further results are available, Susan Cavan is the most likely candidate.

\rule{0pt}{3ex} \textbf{ASSISTANT:} \textbf{Final Answer:} \texttt{Susan Cavan} \\
\bottomrule
\end{tabular}
\caption{Comparative execution traces showing how year filter constraints lead to hallucination in both cases}
\label{tab:merged-execution-traces}
\end{table}

Both examples demonstrate a consistent pattern where the 8B model's use of additional filtering parameters systematically eliminates relevant search results. These parameters restrict searches to specific subsets of documents, which typically lack the comprehensive information found in broader searches. When faced with limited search results, the 8B model resorts to hallucination—generating "Randy Weaver" for the Dutch businessman query and "Susan Cavan" for the production manager query. In contrast, the 4B model's straightforward approach immediately retrieves accurate information from reliable sources, successfully identifying both Rudi Dekkers and Masaki Iizuka. This pattern suggests that overthinking through additional constraints can lead to systematic failures and hallucinated responses in larger models.

\subsection{Jan-nano Configuration Assessment: smolagents vs LangGraph}\label{appendix:a2}

We evaluated Jan-nano across two agentic implementations to assess format generalization capabilities. For smolagent evaluation, we implemented two tools: SearchTool that returns search result links and VisitTool that returns page content, both serving as wrappers around the Serper API with equivalent functionality to the mcp-server-serper.

\begin{table}[H]
\centering
\begin{tabular}{lc}
\toprule
\textbf{Framework} & \textbf{SimpleQA (Full) Accuracy} \\
\midrule
Jan-nano (smolagents CodeAgent) & 76.2\% \\
Jan-nano (LangGraph MCP) & 80.7\% \\
\bottomrule
\end{tabular}
\caption{Jan-nano performance across agentic framework configurations}
\label{tab:framework-comparison}
\end{table}

Jan-nano shows a 4.5 percentage point performance difference between smolagents CodeAgent (76.2\%) and LangGraph MCP (80.7\%). This relatively modest gap demonstrates the model's ability to adapt across different tool calling formats while maintaining competitive performance.

The performance difference reflects template format alignment. Jan-nano was trained with JSON tool call templates, directly matching LangGraph's format, while smolagents CodeAgent requires Python code generation—a format the model was not specifically optimized for. 

\newpage
\thispagestyle{empty}
\null
\newpage
\newpage
\thispagestyle{empty}
\noindent
\vfill 
\begin{center}
  \resizebox{\textwidth}{!}{%
    \begin{minipage}{3.5\textwidth} 
        \lstinputlisting{ascii-art.txt}
    \end{minipage}%
  }
\end{center}

\vfill 

\end{document}